# Zero-shot object prediction using semantic scene knowledge

Rene Grzeszick and Gernot A. Fink
Department of Computer Science, TU Dortmund University, Germany
{rene.grzeszick, gernot.fink}@tu-dortmund.de

## Abstract

*This work focuses on the semantic relations between scenes and objects for visual object recognition. Semantic knowledge can be a powerful source of information especially in scenarios with few or no annotated training samples. These scenarios are referred to as zero-shot or few-shot recognition and often build on visual attributes. Here, instead of relying on various visual attributes, a more direct way is pursued: after recognizing the scene that is depicted in an image, semantic relations between scenes and objects are used for predicting the presence of objects in an unsupervised manner. Most importantly, relations between scenes and objects can easily be obtained from external sources such as large scale text corpora from the web and, therefore, do not require tremendous manual labeling efforts. It will be shown that in cluttered scenes, where visual recognition is difficult, scene knowledge is an important cue for predicting objects.*

## 1. Introduction

Much progress has been made in the field of image classification and object detection, yielding impressive results in terms of visual analysis. Latest results show that up to a thousand categories and more can be learned based on labeled instances [25]. In comparison, it is estimated that humans recognize about 30.000 visual categories and even more sub-categories such as car brands or animal breeds [20]. Human learning is different from machine learning, although it can be based on visual examples, it is also based on external knowledge such as descriptions of entities or the context in which they appear. Recognition systems often omit basic knowledge on a descriptive level. This work will focus on the semantic relations between scenes and objects which can be an important cue for predicting objects. This is especially useful in cluttered scenes where visual recognition may be difficult. Knowing the scene context, which yields a strong prior on what to expect in a given image, the presence of objects in an image will be predicted.

It is known that contextual information can help in the task of recognizing objects [4, 5, 31]. Various forms of context that can improve visual recognition tasks have already been investigated in [5]. Visual context can be obtained in a very local manner such as pixel context or in a global manner by image descriptors like the Gist of a scene [19]. Another form of visual context is the presence, appearance or location of different objects in a scene. External context cues are, for example, of photogrammetric, cultural, geographic or semantic nature [5].

Especially the object level approaches that define context based on the dependencies and co-occurrences of different objects are pursued in several works [4, 9, 27]. In [9] a stacked SVM classifier is applied that uses the maximal detection scores for each object category in an image in order to re-rank the prediction scores. The work presented in [4] uses a hierarchical tree structure in order to model the occurrence of objects as well as a spatial prior. In [27] the detection scores of a specific bounding box are re-ranked based on its position in the scene as well as the relative position of other bounding boxes. More global approaches use image level context definitions for object detection or image parsing [15, 18, 26]. Most of these context definitions follow the same approach: They retrieve a subset of training images which are similar to the given image and transfer the object information [15, 26]. A slightly different approach is pursued in [18], where a Context Forest is trained that learns the relation between a global image descriptor and the objects within this image. This allows to efficiently find related images based on the forest's leaf nodes and then transfer assumptions about object locations or classes. Most of these works have in common that a considerable effort went into training a state-of-the-art detector and the results of various detections are combined in order to obtain a context descriptor. In [4, 9, 18] these were deformable part based models that build on HOG features. Nowadays, Convolutional Neural Networks (CNNs), like very deep CNNs [25] and R-CNNs [10, 22] show state-of-the-art performance in object prediction and object detection respectively. While methods like data augmentation and pre-training have reduced the required number of samples and weakly super-vised annotation schemes lower the required level of detail, still a considerable annotation effort is required to train these models.

A different idea is adding further modalities for context. The most prominent of these modalities is text, for example, image captions or additional tags [14]. These multi-modal approaches allow for answering visual queries [12, 28, 31], the captioning of images or videos [6, 24] and recognition with limited training samples supported by additional linguistic knowledge [13, 23]. Several of these approaches incorporate additional attributes that allow for transferring knowledge without explicitly annotating a specific class or object label [13, 23, 31]. For example, instead of recognizing an animal, visual attributes like its color, whether it has stripes or is shown in water are recognized. These attributes can then be used as additional visual cues.

In [21] a database that focuses on scene attributes is introduced. Each scene is described based on its visual attributes such as natural or manmade. Attributes are predicted independently of each other using one SVM per attribute. Following up on the idea of attribute prediction, it has been shown that a combined prediction of these attributes can be beneficial as they are often correlated. In [11] a neural network is trained that predicts multiple attributes simultaneously and outperforms the traditional per class SVMs on scene attributes.

In [31] a knowledge base system is built that builds on a similar idea and relates scenes with attributes and affordances. It is shown that the association of scenes with attributes and affordances allows for improving the predictions of scenes and their attributes as well as answering visual queries. An even more complex system is presented in the Visual Genome [12] where a complete scene graph of objects, attributes and their relations with each other is presented. This allows not only for predicting attributes and relations, but also for answering complex visual questions. Textual queries are parsed with respect to attributes so that the best matching images can be retrieved.

In [13] attributes, which are associated with a set of images or classes, are used for uncovering unknown classes and describing them in terms of their attributes. For example, an animal with the attributes *black, white* and *stripes* will most likely be a zebra. Each of the attributes is recognized independently and without any knowledge about the actual object classes. The attribute vector is then used in order to infer knowledge about object classes. Such methods with no training samples for given objects are also referred to as zero-shot learning approaches. Similarly, given a very small set of training samples, attributes can be used in order to transfer class labels to unknown images [23].

Attributes for images can either be learned directly via annotated training images or indirectly via additional sources of information such as Wikipedia or WordNet [13, 23]. As annotating images with attributes is tedious, especially, the latter allows for scaling recognizers to a larger number of classes and attributes. Furthermore, it has been shown that these attributes can also be derived in a hierarchical manner, i.e., based on the WordNet tree [23]. An important factor for incorporating such additional linguistic sources is the vast amount of text corpora that are available on the web. Information extraction systems like TextRunner [1] or Reverb [7] allow for analyzing these text sources and uncovering information, like nouns and the relations between them.

This work will show that additional textual information is beneficial for predicting object presence in an image with minimal annotation effort similar to [13, 23, 31]. Here, instead of attributes, a more direct way is proposed exploiting the semantic relations between scenes and objects in a zero-shot approach. The presence of an object is predicted based on two sources: visual knowledge about the scene and the relations between scenes and different objects. For example, a car can hardly be observed in the livingroom or a dining table in the garage. Such knowledge can easily be obtained from additional textual sources using methods like TextRunner [1] or Reverb [7]. As a result the tremendous annotation effort that is required for annotating objects in images is no longer necessary. Similar to existing zero-shot approaches, it requires only a descriptive label, the scene name, and works completely unsupervised with respect to annotations of objects. In the experiments it will be shown that such high level knowledge allows for predicting the presence of objects, especially in very cluttered scenes.

## 2. Method

In the proposed method for object prediction, the relations between scenes and objects which are obtained from text sources are used for modeling top-down knowledge. They replace the visual information that is typically used for object prediction. An overview is given in Fig. 1. The image is solely analyzed on scene level, which requires minimal annotation effort and no visual knowledge about the objects within the scene. More importantly, a text corpus is analyzed with respect to possible scenes and objects, extracting the relations between them and creating a matrix of objects in a scene context. This information is then used in combination with the scene classification in order to predict the presence of an object in an image.

### 2.1. Relations between objects and scenes

Given a large enough set of text, relations extracted from these texts can be assumed to roughly represent relations that are observed in the real world and henceforth may also be observed on images. Rich text corpora can, for example, be obtained by crawling Wikipedia or any other source of textual information from the web [23]. Here, sentences including possible scene or object categories and their relations will be of further interest. In the following extractions based on Reverb [7] are used. Reverb extracts relations and their arguments from a given sentence. Therefore, two steps are performed: First, for each verb $v$ in the sentence, the longest sequence of words $r_v$ is uncovered so that $r_v$ starts at $v$ and satisfies both a syntactical and a lexical constraint. For the lexical constraint, Reverb uses a dictionary of 1.7 million relations. The syntactical constraint is based on the following regular expression:

$$V|VP|VW^*P$$

$$V = \text{verb particle? adv?}$$

$$W = (\text{noun}|\text{adj}|\text{adv}|\text{pron}|\text{det})$$

$$P = (\text{prep}|\text{particle}|\text{inf. marker}) \quad (1)$$

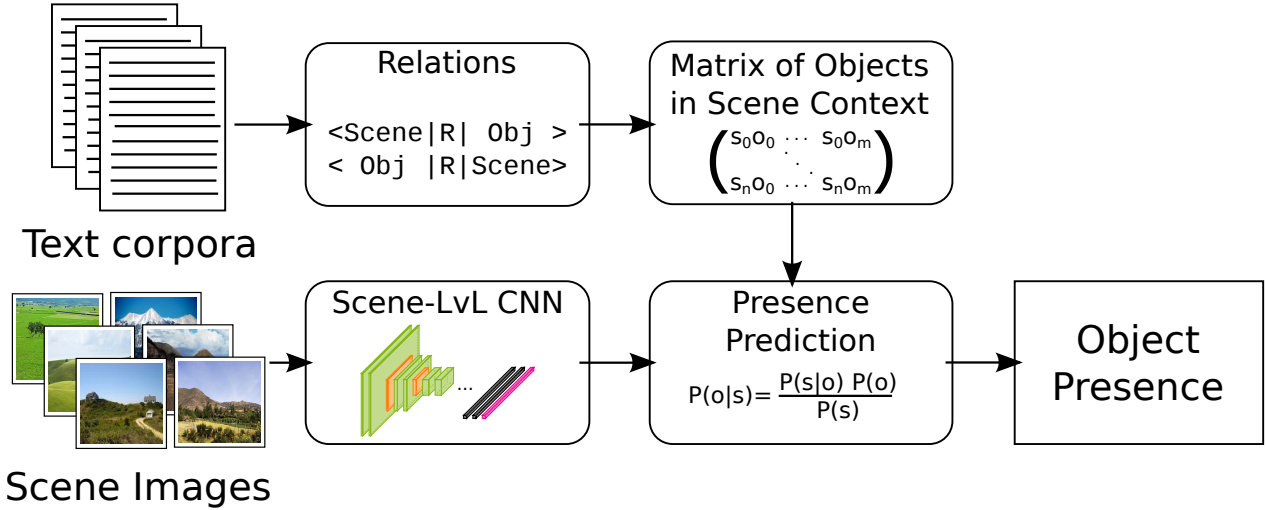

Figure 1. Given external text sources, these are analyzed with respect to possible scenes, objects and their relations, creating a matrix of objects in scene context. For a given set of images, a CNN is trained in order to predict scene labels. This information is then used in order to predict the presence of an object in scene.

Overlapping sets of relations are merged into a single relation. Second, given a relation $r$, the nearest noun left and right of the relation $r$ are extracted. If two nouns can be observed for a relation $r$, this results in a triplet

$$t = (\text{arg1}, \text{r}, \text{arg2}) \,. \quad (2)$$

Simple examples relating scenes and objects may be *'A car drove down the street'* or *'Many persons were on the streets'*. Both relate an entity that is typically used in object detection (*car* or *person*) with a scene label (*street*). In this example, the triplets

$$t_0 = (car, drove\ down, street)\ ,$$
$$t_1 = (person, were\ on, street)$$

would be extracted from the two sentences.

Given a set of relations that were extracted from a text corpus and a vocabulary that defines a set of $S$ scene names and $O$ object names, a matrix $C$ describing the objects in scene context is created. In a general setting a vocabulary could be derived from frequently occurring words in a text corpus, the WordNet tree [17] or just a set of objects and scenes that are of interest and known beforehand. At the index $s, o$ the matrix $C$ contains the number of relations between the respective scene $s$ and object $o$:

$$C_{s,o} = \sum_r n(o,s) + \sum_r n(s,o) \quad \text{with} \quad (3)$$

$$n(i,j) = \#\{t = (i, \cdot, j)\} \quad . \quad (4)$$

Hence, the type of relation is discarded as only the number of relations between an object and a scene will be of further interest. In the experiments, some rare cases of self-similarity were observed, in which a scene and an object name are the same (e.g. a scene in a street may also show the object road/street, among others). In these cases the self-similarity is set to the maximum count observed.

### 2.2. Presence Prediction

The task of presence detection is concerned with the question whether an object can be observed one or more times in a given image. Under the assumption that a large diverse text corpus is representative for the real world, it can further be assumed that the likelihood of an object to occur in a given scene is correlated with the number of textual relations between those entities. Since multiple objects can occur in a single scene image, the presence predictions for different object classes are typically evaluated independently of each other. Due to this multiplicity, the probability $P(o|s)$ of the object $o$ to be shown in the scene $s$ cannot be computed directly using the counts. However, since an image can only depict one scene, $P(s|o)$ can be estimated from the counts. This allows to compute $P(o|s)$ based on Bayes theorem as

$$P(o|s) = \frac{P(s|o)P(o)}{P(s)} \quad \text{with}$$

$$P(s|o) = \frac{C_{s,o}}{\sum_{s'} C_{s',o}} \quad \text{and} \quad P(o) = \frac{\sum_s C_{s,o}}{\sum_{s'} \sum_{o'} C_{s',o'}}.$$
(5)

The prior probability $P(o)$ can be approximated, assuming that one relation count represents the presence of at least one object $o$.

However, for the same reason $P(o|s)$ cannot be derived from the matrix of objects in scene context, the prior probability $P(s)$ for a certain scene cannot be derived. Assume the Matrix $\boldsymbol{C}$ contains $N$ relation counts which relate the presence of at least $N$ objects from $O$ categories to the set of $S$ scenes. Given the one to many relation between a scene and objects, the true count of scenes cannot be recovered. Therefore, $P(s)$ is assumed to be uniformly distributed.

In order to be able to predict an object in a scene where no relations have been previously observed, unobserved events need to be handled. Therefore, the probability of an object $o$ to occur in a scene $s$ is smoothed by

$$P^*(o|s) = (1-\alpha)\, P(o|s) + \alpha\, P(o),$$
(6)

similar to the smoothing of probability distributions for statistical natural language processing (cf. [16]). The process is based on an interpolation factor $\alpha$ which is estimated based on the number of relations with only a single occurrence:

$$\alpha = \frac{\#\{C_{o,s}|C_{o,s}=1\}}{\sum_{o'} \sum_{s'} C_{o',s'}}$$
(7)

Furthermore, the counts are obtained from a text source that is unrelated to the visual tasks so that there is a remaining degree of uncertainty. The matrix representation does also not cover intra-scene variability (i.e. all scenes would have exactly the same relation to objects). In order to model these two issues, $\hat{P}(o|s)$ is sampled by $D$ draws from a normal distribution

$$\hat{P}(o|s) = \frac{1}{D} \sum_1^D n \quad \text{with} \quad n \sim \mathcal{N}(P^*(o|s), \sigma(\boldsymbol{C})).$$
(8)

The variance $\sigma$ is estimated based on the variance within the matrix $\boldsymbol{C}$. In order to estimate the probability of an object in a given image $I$, the presence is then predicted by:

$$P(o|I) = \sum_s P(s|I) \cdot \hat{P}(o|s).$$
(9)

The probability $P(s|I)$ can be predicted by a classifier. Assuming a perfect classification $P(s|I)$ would equal to one for the true scene $s$ and the probability $P(o|I)$ would solely be computed based on the relations between this scene and the objects. However, in practice there will be a distribution over a set of scene labels. This also takes into account the ambiguity between different scenes. In this work, a CNN that is based on a VGG16 network architecture is used for predicting the scene category [25]. The network is pre-trained on ImageNet. It is then adapted to the scene classification using a set of scene images depicting $S$ scenes categories.

Note that the requirement for a single scene label is very easy to fulfill. The annotation effort for a single scene label is much lower than labeling various object classes in an image or even annotating the position of an object in a scene which has to be done for most supervised object detectors. Similar to attribute-based zero-shot learning, it is a descriptive abstraction that does not imply any visual knowledge about the objects within the scene.

## 3. Evaluation

In the following the experimental setup and the evaluation of the proposed object prediction are described. Ideally, the evaluation requires a dataset that offers both scene and object labels. Hence, the different branches of the SUN dataset [29] have been chosen for the evaluation: The SUN2012 Scene and Object Dataset in Pascal VOC format and the SUN2009 Context dataset. Both branches of the SUN dataset show a broad set of different scene and object categories.

**SUN2012 Scene and Object Dataset in Pascal VOC format:** the dataset contains a set of images taken from the SUN image corpus. While the more prominent SUN397 dataset is annotated with 397 different scenes labels, this set contains annotations for an additional $4,919$ different object classes [29]. Of all $16,873$ images, $11,426$ are a subset of the SUN397 dataset for which both annotations, scene and object labels, are available. For the remaining $5,447$ no scene annotations are provided.

**SUN2009 Context:** the dataset contains only about 200 different object categories, of which 107 were used for supervised detection experiments in [4]. The same diversity as for the SUN2012 Scene and Object dataset can be observed with respect to the scene and object categories.

In contrast to traditional object detection tasks, like the Pascal VOC challenge [8], there is a great variability with respect to the objects properties. While some of them are well defined (e.g. cars, person), some others describe regions (sky, road, buildings) or highly deformable objects (river, curtain). Moreover, the annotations in all versions of the SUN dataset are very noisy. Some of them contain descriptive attributes, like *person walking*, *table occluded*, *tennis court outdoor* others mix singular and plural.

In order to relate the scene and object labels with natural text, these descriptive attributes were removed and all objects and scene labels were lemmatized based on the WordNet tree [17]. This leaves $3,390$ unique object labels in 377 different scenes labels. Although the lemmatized scene names may be semantically similar, they may be visually different (i.e. for *tennis court indoor* and *outdoor*). Therefore, the scene labels for all 397 labels will be predicted and the prediction results will be summarized. The objects are then predicted based on the lemmatized names.

### 3.1. Creating a Matrix of Objects in Scene Context

In order to obtain a matrix of objects in scene context, the OpenIE database has been queried. It contains over 5 billion extractions that have been obtained using Reverb on over a billion web pages[1]. Hence, a very diverse dataset that captures the relations between a huge set of nouns has been used. The vocabulary has been defined based on the task of the SUN2012 Scene and Object dataset so that the vocabulary consists of the $S = 377$ scene names and $O = 3390$ object names. All possible combinations of scenes and objects were queried for which a total relation count of $1,375,559$ has been extracted[2]. Note that the distribution of these relation counts is very long tailed, leaving a large set of unobserved events.

### 3.2. Scene prediction

For recognizing objects based on the scene context, the probability of the given image to depict the scene $s$ needs to be computed. In the following, two different setups are evaluated.

**Perfect classifier:** it is assumed that the scene label is known beforehand (i.e. given by a human in the loop) or that training a perfect scene classifier with respect to the annotated scene labels would be possible. In order to simulate this case, $P(s|I)$ is set to 1 for the annotated scene label and to 0 otherwise. Note that these labels might be ambiguous and even human annotators deviate in their decision from the ground truth labels [30].

**Scene-level CNN:** For recognizing scene labels a CNN is evaluated. A VGG16 network architecture has been pretrained on ImageNet. It has then been adapted to the task of scene classification using all scene images from the SUN397 dataset that are not included in the SUN2012

---
[1] For a demo see http://openie.allenie.org
[2] The Matrix of Objects in Scene Context will be made publicly available together with the detailed experimental setup containing the training/test split and the lemmatized annotations at http://patrec.cs.tu-dortmund.de/cms/en/home/Resources/index.html.

Table 1. Recognition rate for the $k$ highest scoring prediction of the scene label using a CNN.

| k-best | Recognition rate |
|---|---|
| 1 | 62.2% |
| 3 | 82.8% |
| 5 | 88.9% |

Scene and Object dataset. The exclusion of the images from the SUN2012 Scene and Object dataset leaves a set of $97,304$ training images. The training images have been augmented using random translations ($0 - 5\%$), flipping (50% chance) and Gaussian noise ($\sigma = 0.02$) in order to achieve a better generalization. In total $500,000$ training images have been created. The learning rate has been set to $\alpha = 0.0001$ using $25,000$ training iterations of batch size $39 (= 10\%$ of the number of classes).

The CNN recognition rates for the $k$ highest scoring predictions on the SUN2012 Scene and Object dataset are shown in Tab. 1. The highest scoring prediction yields an accuracy of $62.2\%$. When considering that the correct result must be within the five highest scoring predictions an accuracy of $88.9\%$ is achieved. This emphasizes that the probabilistic assignment to a set of scene categories based on the CNNs predictions is a meaningful input for the proposed object prediction.

### 3.3. Object prediction

In the following experiments, the presence detection is evaluated. Note that only scene labels were used for training the CNN so that it is completely unsupervised with respect to object occurrences.

**SUN 2012 Scene and Object dataset**

For the SUN2012 Scene and Object dataset, the most frequently occurring 100 up to all 3390 objects categories from the dataset were considered, some of them being comparably rare. The accuracy of the top $k$ predictions is evaluated. Note that multiple objects can occur in a single image and, therefore, the accuracy of all predictions is evaluated (i.e. for $k = 2$, both predictions are compared to the ground truth so that each one can be a correct or false prediction). Only images with at least $k$ annotated objects were used for the evaluation. Figure 2 shows the results when simulating a perfect classifier based on the annotated scene labels and Fig. 3 shows the results when predicting $P(s|I)$ using the CNN. For both experiments the sampling parameter $D$ has been set to 10.

Interestingly, the simulation of a perfect scene classifier does not yield superior results compared to those achieved by predicting the scene label using the CNN. The reason for this is two fold: scenes are often ambiguous (cf. [30]) and the prediction using the CNN computes a probabilistic assignment to a set of scenes. For example, a scene depicting a *cathedral*, *church* or *chapel* may not only be visually similar, but they are also similar on a semantic level. As the distribution of objects in scene context is obtained from a very general external text source, it does not accurately match the ground truth distribution that can be observed in the dataset. Hence, a mixture of many scenes is more robust.

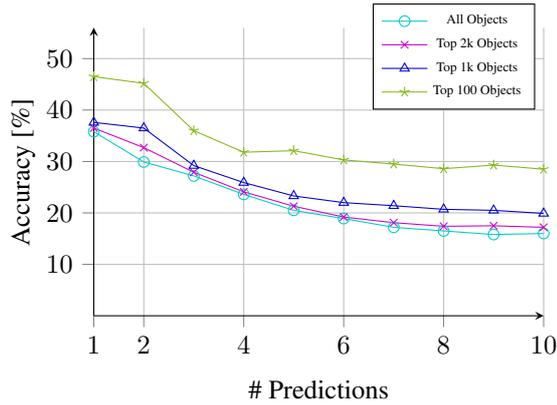

Figure 2. Accuracy for the top k object predictions on the SUN2012 Scene and Object Dataset. The scene label is predicted by simulating a perfect classifier. Hence, the object prediction is only based on the number of relations between the annotated scene and the set of objects.

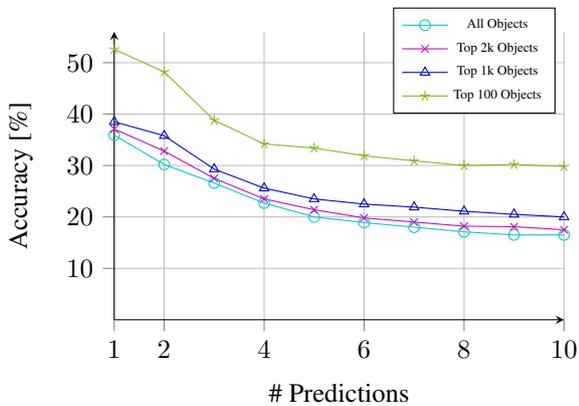

Figure 3. Accuracy for the top k object predictions on the SUN2012 Scene and Object Dataset. The scene labels are predicted using a CNN and the relations between the predicted scenes and the set of objects is used for predicting the object presence.

The results in Fig. 3 also show that even without any knowledge about the visual appearance of an object the highest ranking object predictions have a precision of up to $52.6\%$ when considering a set of 100 objects and $35.9\%$ when considering as many as $3,390$ different object categories. However, as mentioned before, some of the objects describing regional objects tend to be very general an can safely be assumed to occur in most scenes.

Exemplary results showing the five highest scoring object predictions for a given image are shown in Fig. 4. It can be seen that although a large set of objects is annotated in the SUN dataset, the annotations are noisy and not at all complete. Some of the predictions that cannot be found in the ground truth annotations might be deemed as correct. Predictions that are not shown in the image, are often at least plausible guesses what else could be found in the scene. For example, people might be related to a highway or street scene and a sofa might be related to an indoor scene. The third example in the top row is a typical case where the highest scoring prediction of the CNN is incorrect. However, the prediction of livingroom instead of bedroom is not only visually but also semantically related. The wall on the left with the TV could also easily be placed in a livingroom. Furthermore, due to the probabilistic assignment to a set of scenes, the object class bed is still the fifth best scoring prediction although no relation between livingroom and bed has been found in the external text sources. The example of a bathroom in the bottom row shows a typical example of ambiguity in natural language as well as in the provided annotations.

In order to provide a more detailed analysis, different sets of object categories are evaluated based on the VOC mean average precision (mAP) criterion [8]. The results for the 20 to 100 most common objects in the dataset is given in Tab. 2. The mAP of predicting an object by chance is indicating how frequently these objects occur in the dataset. For comparison, the ground truth distribution of all scenes and objects, which has been observed in the dataset, has been evaluated using the same model. This can be seen as an indication of an upper bound for the performance that could be obtained by solely using the proposed model of scene and object relations. It can clearly be observed that the relations obtained from the text sources do not model the distribution in the dataset perfectly. The proposed method shows promising results given the fact that the relations are obtained from arbitrary websites. The results also show that there is potential for improvement if the relation can be estimated in a more accurate manner.

### SUN2009 Context

In order to emphasize the difficulty of detecting objects with a huge variability, as depicted in the SUN dataset, the approach has also been evaluated on the 107 object categories of the SUN2009 Context dataset. On this dataset, different object detectors based on deformable part based models were trained in a fully supervised manner and evaluated in [4]. The presence detection of the proposed zero-shot method after predicting a scene label for each of the scenes using the CNN is compared to the supervised object detectors of [4].

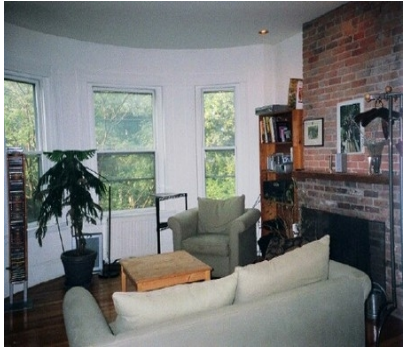 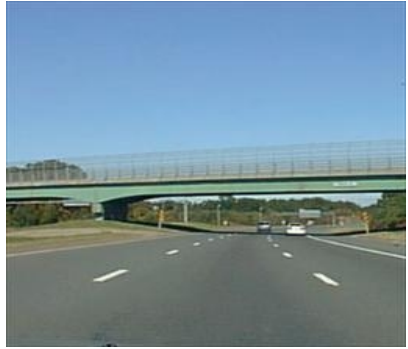 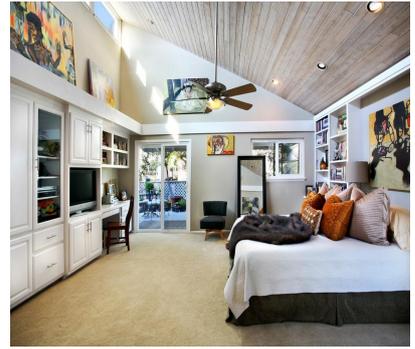

*house*, television, fireplace
floor, sofa
Livingroom — Livingroom

people, bridge, *vehicle*
road, car
Highway — Highway

sofa, television, *house*
floor, bed
Bedroom — Livingroom

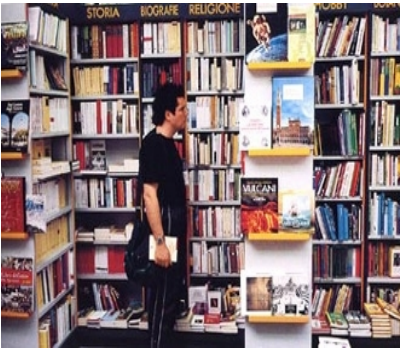 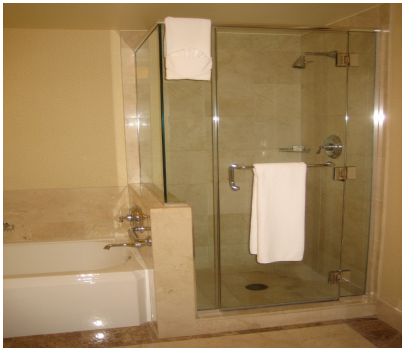 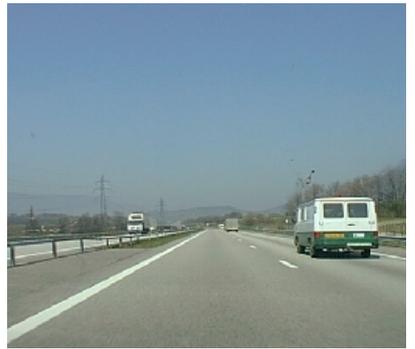

information, *people*, floor
text, book
Bookstore — Bookstore

shower, *bath*, floor
*tub*, bathtub
Bathroom — Bathroom

parking, people, *vehicle*
road, *car*
Highway — Highway

Figure 4. Further exemplary results showing the five highest scoring object predictions: (green) correct (red) wrong (red & italic) wrong according to annotations, but can be seen in the image. In the bottom row: (left) Annotation (right) highest scoring CNN prediction.

Table 2. Mean average precision for different sets of objects on the SUN2012 dataset in Pascal format. The presence predictions are based on the number of relations between the scenes and the objects. ($1^{st}$ col.) Simulation of a perfect classifier using the ground truth scene labels ($2^{nd}$ col.) Scene labels are predicted using a CNN. For comparison the results of a prediction by chance ($3^{rd}$ col.) and the results of the proposed method when using the true number of scene-object relations derived from the dataset ($4^{th}$ col.) are also shown.

| Objects | Perfect Classifier mean AP [%] | Scene-level CNN mean AP [%] | Chance mean AP [%] | GT Distribution mean AP [%] |
|---|---|---|---|---|
| Top 20 | 34.7 | 38.5 | 21.6 | 54.1 |
| Top 40 | 29.5 | 33.1 | 13.9 | 47.1 |
| Top 60 | 24.2 | 27.3 | 10.3 | 43.2 |
| Top 80 | 21.5 | 24.2 | 8.5 | 39.2 |
| Top 100 | 19.4 | 22.1 | 7.2 | 35.9 |

The results are shown in Table 3. As the original evaluation protocol contained three classes that were filtered by the stemming (*bottles*, *stones* and *rocks*) and can therefore not be recognized by the proposed method, the results for all 107 classes as well as the results for the remaining 104 classes after the stemming are displayed. It is surprising that a model that is solely based on scene-level predictions can achieve comparable results to deformable part based models that are trained completely supervised. Only with additional contextual information the part based models are able to outperform the proposed approach. This clearly shows the requirement for contextual information, especially since the since visual information in cluttered scenes may be limited.

Table 3. Mean average precision for the object presence in the SUN2009 Context dataset. (*) The CNN predicts one of the 397 scenes from the SUN397 dataset without any knowledge about the objects.

| Method | Annotations | # Objects | m AP[%] |
|---|---|---|---|
| Part based Models [9] | Cropped objects | 107 | 17.9 |
| PbM + Tree Context [4] | Cropped objects | 107 | 26.1 |
| PbM + Context SVM [4] | Cropped objects | 107 | 23.8 |
| Scene-level CNN + Objects in Context | - (*) | 107 | 19.1 |
| Scene-level CNN + Objects in Context | - (*) | 104 | 19.8 |

Note that although there is no evaluation of R-CNNs on this task, they have surpassed deformable part based models as the state-of-the-art in object detection [10]. It can be assumed that they outperform the part based models on this task as well. Nevertheless, deformable part based models are powerful object detectors and it is interesting that an unsupervised approach can achieve similar results. This shows that the relations between scenes and objects provide important cues for object prediction.

## 4. Conclusion

In this work a novel approach for predicting object presence in an image has been presented. The method works in a zero-shot manner and only relies on scene level annotations from which a probability for an object's presence is derived. The probability is based on the relations between scenes and objects that were obtained from additional text corpora. As a result the proposed method is completely unsupervised with respect to objects and allows for predicting objects without any visual information about them. In the experiments it has been shown that it is possible to predict the occurrences for as many as 3390 objects. On tasks that are very difficult for visual classifiers, such as cluttered scenes with not very well structured objects, the approach yields similar performance to visual object detectors that were trained in a fully supervised manner.

## Acknowledgements

This work has been supported by the German Research Foundation (DFG) within project Fi799/9-1. The authors would like to thank Kristian Kersting for his helpful comments and discussions.

## 5. Supplementary Material

This supplementary material of the submission contains additional evaluations on the SUN dataset as well as evaluations on additional datasets.

### 5.1. SUN2012 Scene and Objects - VOC Objects

The mean average precision for the presence of the 20 VOC objects classes in the SUN2012 Scene and Object dataset is shown in Tab. 4. The presence predictions are based on the number of relations between the scenes and the objects: (left) The scene label, (right) the predicted scene labels using a CNN. For comparison the prediction by chance is shown and the results using the ground truth distribution of objects in the SUN scenes using the same model. The ground truth distribution can be seen as an indicator for an upper baseline.

### 5.2. VOC 2007 dataset - Presence prediction

The Pascal VOC 2007 dataset contains 20 well defined object classes in less cluttered scenes than in SUN dataset. The limitations of the proposed object prediction can be observed on the this dataset. Here, the approach is generalized to a completely different dataset using the same scene and object categories as gathered for the SUN dataset. The presence of the 20 classes of the VOC benchmark is predicted with the results shown in Table 5. The mean average precision for the presence for each of the 20 VOC objects classes is shown in Tab. 6. For predicting the classes, some of the labels were mapped to the best matching synonym ('*pottedplant*'-'*plant*', '*diningtable*'-'*table*' '*aeroplane*'-'*airplane*' and '*tvmonitor*'-'*television*'). Note that when generalizing to this dataset, it is assumed that the scenes in the SUN dataset and their characteristics are also representative for the scenes found in the VOC dataset. Furthermore, the VOC task is especially designed for visual object detection and the classes are not as ambiguous as the ones in the SUN dataset. They can therefore be learned more efficiently by a visual classifier. However, the proposed approach is still able to make an 'educated guess'.

### 5.3. VOC 2007 - Object Detection

Given a more informed setting, where a set of object detectors has been trained and evaluated, the obtained presence probability can also be used for modeling a context descriptor. Following the approach of the Context SVM [9], the output of an object detector can be re-scored. Here, a Scene-Context SVM is trained. Let $(B, y)$ be the detection window of an object detector, i.e. an R-CNN [22], described by a bounding box $B$ and a score $y$. Then, for each detected bounding box a context descriptor is derived:

$$g = (y, P(o_1|I), .., P(o_n|I)) \qquad (10)$$

Table 6. Mean average precision for the presence detection of the 20 VOC objects classes on the VOC 2007 dataset. The predictions are based on the Scene Label CNN + Objects in Context.

| Objects | MaP | | |
|---|---|---|---|
| Person | 49.21 | Bus | 13.14 |
| Bird | 12.84 | Car | 51.61 |
| Cat | 8.34 | Motorbike | 6.38 |
| Cow | 33.73 | Train | 5.35 |
| Dog | 13.70 | Bottle | 16.31 |
| Horse | 22.64 | Chair | 47.36 |
| Sheep | 25.91 | Diningtable | 44.50 |
| Aeroplane | 80.55 | Pottedplant | 7.92 |
| Bicycle | 12.90 | Sofa | 60.10 |
| Boat | 43.33 | Tvmonitor | 12.94 |
| | | VOC Mean | 28.44 |

Table 7. Mean average precision for the detection using context rescoring for the 20 objects in the VOC2007 dataset.

| Method | m AP[%] |
|---|---|
| R-CNNs [22] | 69.6 |
| + Context SVM [9] | 69.1 |
| + Scene-Context SVM | **69.8** |
| Upper Bound | 80.69 |

The location and object co-occurrence from the re-scoring in [9] are replaced by the scene priors. One SVM is trained for each class on the true positive and false positive detections using a linear kernel and Platt's scaling.

Although the presence detection on the VOC dataset appears to be difficult in a unsupervised manner, it can be shown that the object prediction scores are still a powerful context cue. The results for an R-CNN object detector using a Region Proposal Network with 300 proposal have been rescored by the proposed Scene-Context SVM and are shown in Tab. 5. The mean average precision for the detection 20 VOC objects classes on the VOC2007 dataset is shown in Tab. 8. For comparison, a Context SVM as introduced in [9] has been evaluated. Both approaches were trained on the VOC2007 *trainval* set. Although the theoretical upper bound allows for some improvement, the low performance of the Context SVM emphasizes the difficulty of re-ranking the CNN's predictions. This is mainly due to the fact that the network's confidence in its predictions is typically quite high. Still the proposed Scene-Context SVM that integrates a strong prior is able to slightly improve the results.

The Training Parameters for the Scene-Context SVM were: *Linear Kernel, No shrinking heuristic, C=1, Train Tolerance=0.01*

Table 4. Mean average precision for the 20 VOC objects classes on the SUN Pascalformat dataset.

| Objects | Scene Label mean AP [%] | Scene Label CNN mean AP [%] | Chance mean AP [%] | GT Distribution mean AP [%] |
|---|---|---|---|---|
| Person | **19.6** | 18.7 | 17.6 | 45.0 |
| Bird | 0.2 | 0.2 | 0.1 | 3.6 |
| Cat | 0.2 | 0.1 | 0.1 | 0.4 |
| Cow | 28.0 | **37.2** | 0.1 | 28.0 |
| Dog | **5.7** | 2.1 | 0.4 | 28.2 |
| Horse | **2.7** | 2.2 | 0.1 | 39.1 |
| Sheep | **20.7** | 5.6 | 0.1 | 7.7 |
| Airplane | 11.9 | **14.9** | 0.6 | 81.7 |
| Bicycle | **1.6** | 1.2 | 0.4 | 11.8 |
| Boat | 8.2 | **9.5** | 1.2 | 18.6 |
| Bus | 2.4 | **6.0** | 0.4 | 4.7 |
| Car | 48.6 | **59.1** | 8.1 | 66.1 |
| Motorbike | 1.5 | **2.6** | 0.3 | 4.3 |
| Train | 0.2 | 0.2 | 0.2 | 88.8 |
| Bottle | 19.1 | **19.3** | 5.2 | 23.1 |
| Chair | 37.8 | **39.3** | 18.1 | 59.4 |
| Table | **36.2** | **36.2** | 13.9 | 53.9 |
| Plant | 16.2 | **17.9** | 14.6 | 26.2 |
| Sofa | 32.5 | **49.6** | 4.5 | 47.2 |
| Television | 7.6 | **7.7** | 2.9 | 11.3 |
| VOC Mean | 15.0 | **16.5** | 4.5 | 32.3 |

Table 5. Mean average precision for the presence prediction of the 20 objects in the VOC2007 dataset. (*) The CNN predicts one of the 397 scenes from the SUN397 dataset without any knowledge about the VOC dataset.

| Method | Annotations | m AP[%] |
|---|---|---|
| CNN [3] | Tagged Objects | 82.4 |
| BoF - Fisher Kernel [2] | Tagged Objects | 61.7 |
| BoF [2] | Tagged Objects | 55.3 |
| SL CNN + Obj. in Context | - (*) | 28.4 |
| Chance | - | 8.1 |

Table 8. Mean average precision for the detection of the 20 VOC objects classes on the VOC 2007 dataset.

| Objects | R-CNN [10] | + Context SVM [9] | + Scene-Context SVM | Upper Bound |
|---|---|---|---|---|
| Person | 75.73 | 75.56 | **75.76** | 80.74 |
| Bird | **69.42** | 68.94 | 69.13 | 80.71 |
| Cat | **85.03** | 84.27 | 85.01 | 90.61 |
| Cow | 76.48 | 75.48 | **77.20** | 90.17 |
| Dog | 81.11 | **81.33** | 81.26 | 90.63 |
| Horse | **83.58** | 83.22 | 83.32 | 90.46 |
| Sheep | 70.48 | 64.40 | **70.94** | 81.82 |
| Aeroplane | **68.74** | 68.48 | 68.72 | 71.89 |
| Bicycle | 78.33 | 78.41 | **78.36** | 81.54 |
| Boat | 54.55 | 54.92 | **55.62** | 72.36 |
| Bus | 79.62 | 77.80 | **79.69** | 90.72 |
| Car | **79.79** | 79.51 | 79.76 | 80.84 |
| Motorbike | **76.21** | 76.16 | 75.98 | 81.27 |
| Train | **77.88** | 78.42 | 77.38 | 81.47 |
| Bottle | 49.70 | 49.42 | **49.92** | 62.68 |
| Chair | **50.98** | 50.30 | 50.96 | 71.54 |
| Diningtable | 65.34 | 62.91 | **65.64** | 81.56 |
| Pottedplant | **38.40** | 37.36 | 38.34 | 62.81 |
| Sofa | 65.80 | 64.40 | **66.14** | 88.52 |
| Tvmonitor | 65.70 | 63.38 | **65.76** | 81.43 |
| VOC Mean | 69.64 | 69.07 | **69.75** | 80.69 |